# ASSS: A Differentiable Adversarial Framework for Task-Aware Data Reduction

Jiacheng Lyu[1, *]
*School of Computer Science and Technology, Xidian University*, Xi'an, China
*School of Mathematical and Computer Science, Heriot-Watt University*, Edinburgh, UK
25032100013@stu.xidian.edu.cn

Bihua Bao[1]
*School of Economics and Management, Xdian University*, Xi'an, China
*Pamplin College of Business, Virginia Polytechnic Institute and State University*, Blacksburg, US
25061300025@stu.xidian.edu.cn

Shiyun Yan
*School of Information Engineering, Wuhan University of Technology*, Wuhan, China
386781@whut.edu.cn

***Abstract*—Massive datasets often contain redundancy that inflates computational costs without improving generalization. Existing data reduction methods are typically task-agnostic, discarding informative boundary samples and yielding suboptimal performance. We propose Adversarial Soft-Selection Subsampling (ASSS), a differentiable framework that casts data reduction as a minimax game between a learnable selector and a task network. Using Gumbel-Softmax relaxation, ASSS enables end-to-end gradient flow and is theoretically grounded in the information bottleneck principle. Experiments on multiple benchmarks show that ASSS achieves a performance retention rate (PRR) of 98.9% while using only 30% of the data, significantly outperforming random sampling, K-means, and gradient-based methods. Visualizations confirm that ASSS preferentially retains samples near decision boundaries. The framework is scalable, fully differentiable, and easily integrated into existing training pipelines. This work introduces a new paradigm for task-aware data reduction that directly optimizes subset selection for the downstream objective, offering a principled and practical solution to the scalability challenges in modern deep learning.**

## I. Introduction

The exponential growth of data across domains ranging from high frequency financial trading records and large scale medical imaging archives to massive sensor networks in cyber-physical systems has fundamentally promoted the rise of modern data-driven machine learning [1]. While deep neural networks have achieved unprecedented performance in tasks like image classification and time-series forecasting, scaling their training to massive datasets demands prohibitive computational resources [2]. More critically, these collections frequently harbor substantial redundancy and noise that inflate costs while degrading generalization through overfitting to spurious correlations, ultimately yielding diminishing performance returns [3]. To alleviate these burdens, data reduction approaches like subsampling are developed [4]. By selecting informative subsets of data, these methods overcome the scalability barriers imposed by massive, high-dimensional datasets while maintains or even enhances the model performance [5].

Traditional data reduction techniques, such as random sampling [6], clustering-based core-set selection [7], and the nearest-neighbor thinning [8], have long served as important pre-processing steps to mitigate these scalability bottlenecks. However, crucially, these methods are operated under a fundamentally task-agnostic assumption, selecting subsets based solely on geometric properties like cluster centroids [7] and local density [8] or statistical heuristics [9], without considering the downstream learning objective. Either do these methods ignores the inherent heterogeneity of sample information, or omit critical patterns or retain redundant and noisy instances, leading to suboptimal and unstable model performance [10]. As a result, they frequently discard samples lying near decision boundaries or encoding rare but highly discriminative patterns, which are precisely the instances most valuable for accurate predictions in a given task [11]. This decoupling between data selection and task optimization results in suboptimal performance retention rates (PRR), particularly acute in high-dimensional data and in imbalanced real-world scenarios [12]. In high-dimensional data, the complex, non-linear, and subtle feature relationships make simple geometric distances or global statistical heuristics inadequate for identifying task-relevant patterns for predictive performance [13].

In this paper, we introduce the Adversarial Soft-Selection Subsampling (ASSS) framework as a paradigm shift: reframing data subsampling as a dynamic, fully learnable, task-aware component integrated end-to-end into the model training pipeline. The originality of the work are mainly threefold as following:

- Adversarial Game Formulation: We cast subsampling as a minimax game between a differentiable Selector and a Task network, jointly optimizing for maximum predictive utility under sparsity constraints.
- End-to-End Differentiability: Via Gumbel-Softmax reparameterization, ASSS enables fully gradient-based, differentiable sample selection seamlessly integrated into standard training pipelines.
- Information-Theoretic Principle: ASSS implicitly optimizes an information bottleneck, maximizing mutual information between selected subsets and targets while minimizing complexity, providing principled theoretical justification.

The paper is structured as follows: Section II reviews related work in data reduction and adversarial methods. Section III describes the ASSS method in detail. Section IV provides experimental result with analysis and conclusion is given in Section V.

## II. Related Work

### A. Data Reduction and Core-set Selection

Early approaches to handling large-scale data focused on identifying representative subsets. Random subsampling is computationally trivial but ignores sample heterogeneity. Clustering-based methods like K-means prototypes [14], and nearest-neighbor thinning [8], aim to preserve geometric diversity or reduce local density, yet remain task-agnostic and often discard discriminative boundary samples critical for classification. Gradient-based importance sampling [15] and core-set constructions [16], such as those minimizing maximum discrepancy, require multiple passes over the data and lack end-to-end differentiability with respect to the downstream objective. Recent works such as GLISTER [17], and Grad-Match [18] attempt to align selection with model gradients, but still operate as separate preprocessing steps rather than jointly optimized components.

### B. Adversarial and Differentiable Subset Selection

Generative Adversarial Networks (GANs) have inspired minimax frameworks for data-related tasks [19], yet existing differentiable subset selection methods employing Gumbel-Softmax [20] or concrete relaxations typically target architecture search or synthetic data generation, such as dataset distillation [21], rather than instance-level selection from real data. While learning-to-learn paradigms such as active learning policies focus on sequential, label-cost optimization rather than training efficiency [22], and recent core-set advances lack direct task-aware adversarial feedback, ASSS repurposes the adversarial structure for end-to-end data efficiency: instead of generating synthetic samples, the selector network G learns to assign continuous importance weights to existing data points via Gumbel-Softmax reparameterization, enabling gradients from the task loss to flow back through the sampling process. This allows G to adaptively weight real instances to preserve interpretability and fidelity while "challenging" the task network C to achieve high performance from the sparsified subset, effectively adapting selection probabilities to sample features for efficient training.

### C. Information Bottleneck

The information bottleneck (IB) framework formalizes the trade-off between compression and prediction accuracy [23]. While IB has been successfully applied to feature selection and representation learning, its application to instance-level subsampling remains largely theoretical [24]. Our work provides both a practical variational approximation and the first large-scale empirical validation of IB in the context of differentiable data reduction. ASSS bridges these lines of research by casting subsampling as a differentiable adversarial game that directly optimizes an IB objective, while introducing practical efficiency mechanisms that enable scaling to million-sample regimes.

## III. Methodology

### A. Problem Formulation and Notation

Let $D=\{(x_i, y_i)\}_{i=1}^{N}$ denote a large-scale training dataset, where $x_i \in \mathrm{R}^d$ is a feature vector of the $i$-th sample and $y_i \in \{1, ..., K\}$ is its corresponding label. Our aim is to identify a compact subset $J \subset \{1, ..., N\}$ with $|J| \approx rN$ where $0 < r < 1$ is the target compression ratio such that a model trained on $\{(x_i, y_i)\}_{i \in J}$ achieves performance comparable to that obtained using the full dataset $D$.

We introduce a binary selection indicator $z_i \in \{0, 1\}$ for each sample, where $z_i = 1$ indicates retention. The selection policy is parameterized by a selector network $G_\phi$ that output a probability $p_i = \sigma(G_\phi(x_i))$ . As direct optimization over discrete $z_i$ is intractable, we relax the selection process into continuous, differentiable form (detailed in *III.C*). This formulation naturally leads to a bi-level optimization problem:

$$\min_{\phi} \mathrm{L}_G(\phi, \theta^*(\phi)) \;\; s.t. \;\; \theta^*(\phi) = arg \min_{\theta} \mathrm{L}_C(\theta, \phi) \tag{1}$$

where $\mathrm{L}_C$ is the task loss (e.g., cross-entropy) computed on the weighted subset, and $\mathrm{L}_G$ is the selector's composite objective that balances task fidelity and sparsity.

### B. Information Bottleneck Principle

We base this objective on the Information Bottleneck (IB) principle, which aims to maximize the mutual information between the selected subset and the label while minimizing the mutual information between the subset and the original features:

$$\max_{p(z|x)} I(Z; Y) - \beta I(Z; X) \tag{2}$$

where $Z$ denotes the random variable that corresponds to the selection indicators and $\beta > 0$ is the Lagrange multiplier controlling the compression-prediction trade-off.

Exact computation of $I(Z; Y)$ and $I(Z; X)$ is intractable for high dimensional data. We therefore derive a tractable variational upper bound. Let $C_\theta$ serve as a variational decoder approximating the conditional distribution $q_\theta(y|z)$. The first term admits the lower bound

$$I(Z; Y) \geq \mathrm{E}_{p(z,x,y)}[\log q_\theta(y|z)] + H(Y) \approx -\mathrm{L}_C(\theta, \phi) + const \tag{3}$$

where $\mathrm{L}_C$ is the cross-entropy loss. The second term $I(Z; X)$ is approximated by the expected selection rate $\mathrm{E}[z_i]$ (sparsity penalty). Substituting these approximations yields the selector loss $\mathrm{L}_G(\phi, \theta) = \mathrm{L}_C(\theta, \phi) + \lambda \mathrm{E}[z_i] - \gamma \mathrm{H}(z)$where $\lambda$ corresponds to $\beta$ and $\gamma$ is a small entropy regularization coefficient. The entropy term $\mathrm{H}(z) = -\mathrm{E}[\, z_i \log z_i + (1 - z_i) \log(1 - z_i) \,]$ in the selector loss prevents mode collapse to trivial solution by encouraging the selection probabilities to be close to 0.5. To quantify its effect, we measured the diversity of selected subsets with and without this term on a validation toy set. With $\gamma = 0.01$, the average pairwise Jaccard distance among five independent runs was $0.7386 \pm 0.1648$, while without entropy regularization ($\gamma = 0$), the diversity dropped to $0.4540 \pm 0.2800$, confirming that the term indeed promotes varied selections across runs. This diversity is crucial for exploring different informative samples during the adversarial game.

### C. Differentiable Adversarial Training and Subset Selection

The core of ASSS is a two-player minimax game between the selector network $G_\phi$ and the task network $C_\theta$:

- $G_\phi$ learns to assign continuous importance weights $p_i$ that maximize downstream task performance while enforcing sparsity.
- $C_\theta$ learns to extract maximal predictive signal from the sparsely weighted subset produced by $G_\phi$.

To overcome the non-differentiability of discrete sampling, we employ the binary Gumbel-Softmax relaxation. For each sample, the selector outputs a logit $s_i = G_\phi(x_i, g_i)$ where $g_i$ is the gradient proxy of the sample $x_i$. The relaxed selection weight is:

$$z_i = \frac{exp((s_i + n_i)/\tau)}{exp((s_i + n_i)/\tau) + exp((-s_i + n'_i)/\tau)} \qquad (4)$$

where $n_i$ and $n'_i$ are independent and identically distributed Gumble noise samples such that $n_i = -\log(-\log u_i)$, and $u_i \sim Uniform(0, 1)$, and $\tau > 0$ is the temperature parameter. As $\tau \to 0$, $z_i$ converges to a hard binary decision and during training we anneal $\tau$ exponentially from an initial value (e.g. 1.0) to 0.1.

The task loss is computed on the weighted samples:

$$L_C(\theta, \phi) = -\frac{1}{B}\sum_{i=1}^{B} z_i \log p_\theta(y_i|x_i) \qquad (5)$$

where $B$ is the batch size. As discussed earlier, the selector minimizes the composite objective

$$L_G(\phi, \theta) = L_C(\theta, \phi) + \lambda E[z_i] - \gamma H(z). \qquad (6)$$

Through leveraging Gumbel-Softmax reparameterization for gradient flow, both networks are optimized end-to-end via alternating TTUR (Two-Time-Scale Update Rule) updates with elevated learning rates for the task network. The entropy term $H(z)$ promotes diversity, whereas the sparsity penalty $\lambda E[z_i]$ directly controls compression.

The minimax optimization between selector $G_\phi$ and task network $C_\theta$ follows a two-time-scale update rule (TTUR). Let $\eta_G$and $\eta_C$be the learning rates of the two players, with $\eta_G/\eta_C \approx 4$ to ensure that the task network adapts quickly to the current selection. Under this setting, the game converges to a local Nash equilibrium provided that both losses are smooth and the step sizes satisfy the conditions derived in [25]. In practice, we observe that the selector loss stabilizes after 15 epochs, and the task accuracy on a fixed validation set stops increasing, indicating practical convergence. Fig. 3 illustrates the loss trajectories during adversarial training. The loss decreases rapidly from 0.54 to near zero within 15 epochs, while validation accuracy rises from 0.82 to 0.87, confirming that the minimax game converges efficiently under the TTUR schedule.

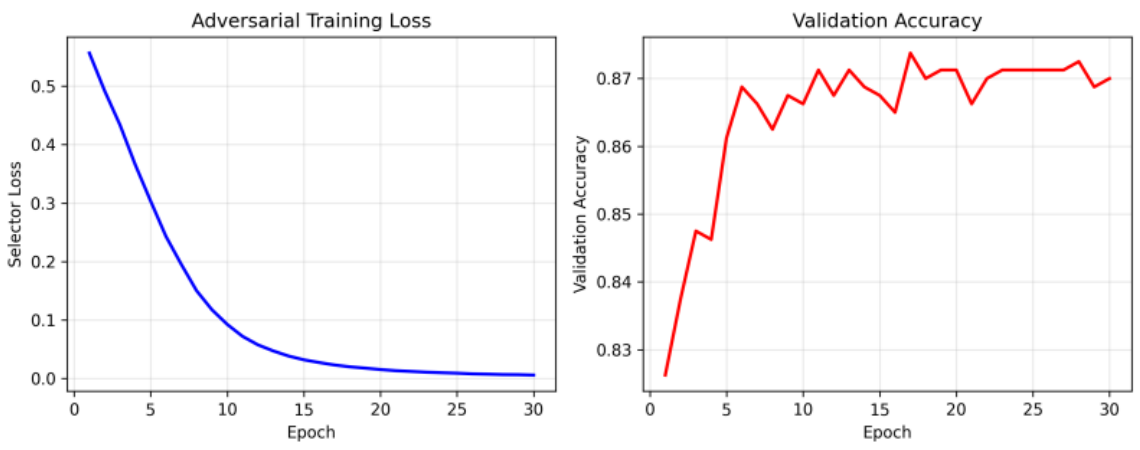

Fig. 1. Adversarial Training Convergance.

### D. Overview of the Training Pipeline

To translate the differentiable adversarial core into a practical and scalable solution for large-scale datasets, we introduce streamlined five-stage training pipeline as is illustrated in Fig. 2:

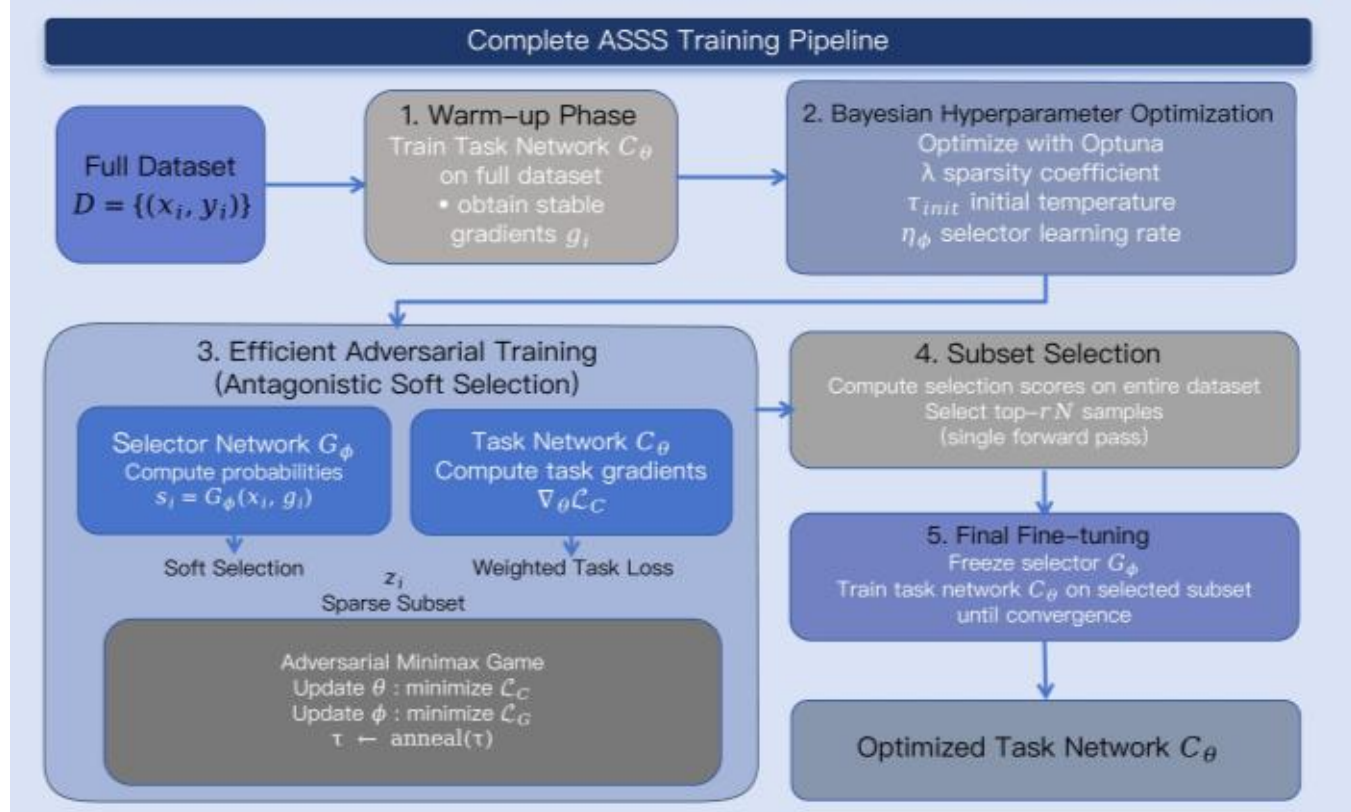

Fig. 2. ASSS efficient five-stage training pipeline.

*1) Stage 1 (Warm-up):* A standard task network $C_\theta$ will be trained for a few epochs on the full dataset using conventional crosss-entropy loss to generate stable per-sample gradient proxies $g_i$ that capture each instance's immediate contribution to thr current decision boundary.

*2) Stage 2 (Bayesian Hyperparameter Optimization):* We automatically search for the optimal sparsity coefficient $\lambda$, initial temperature $\tau_{init}$, and selector learning rate $\eta$using a small validation split (10% of the training data) and Optuna's Gaussian-process surrogate, which can ensure the subsequent adversarial phase begins from a well-tuned configuration without manual trial-and-error.

*3) Stage 3 (Efficient Adversarial Training):* The selector network $G_\phi$ and task network $C_\theta$ are engaged in the minimax game described in III.C. Gradient proxies are concatenated to the selector input, while Gumbel-Softmax relaxation enables differentiable soft selection, then temperature annealing and two-time-scale updates (TTUR) are designed to maintain stability. Only the last layer of $C_\theta$ is actively updated during this phase, keeping computation lightweight.

*4) Stage 4 (Subset Selection):* After adversarial training, selection probabilities are computed once over the entire dataset in a single forward pass, and the top-$rN$ samples are selected based on the learned policy.

*5) Stage 5 (Final Fine-tuning):* The selector is frozen then the complete task network is refined on the selected subset until convergence.

The purpose of this pipeline is to provide the selector with informative initialization signals and well-tuned hyper-parameters ensuring rapid and stable convergence of the core adversarial optimization, while confining computationally intensive minimax interaction to a short, focused phase. Through these stages, the framework avoids mode collapse, maintains end-to-end differentiability, and naturally discovers samples that are maximally informative for the downstream objective. The final fine-tuning step then allows the task network to exploit its full expressive capacity on the curated subset, yielding a model both compact and highly performant.

In summary, ASSS reframes subsampling as a fully differentiable adversarial game that directly optimizes an information-bottleneck objective. It combines gradient surrogate initialization and Bayesian hyperparameter search to significantly accelerate convergence speed and maintain strict task awareness throughout the selection process. All these advances make ASSS transcend static geometric heuristics by learning what truly matter for a given task, prioritizing boundary configurations and rare discriminative patterns while automatically suppressing redundancy and noise, thus providing a principled, scalable, and effective solution for reduced data learning.

## IV. Experiments and Results

### A. Experimental Setup and Network Initialization

To evaluate the effectiveness of ASSS, we conduct experiments on several widely used benchmark datasets form UCI Machine Learning Repository [26] and a synthetic dataset, as is listed in TABLE I. . These datasets vary in size, dimensionality and class distribution, providing diverse scenarios. Due to computational resource constraints, experiments on the HIGGS and KDD Cup 1999 datasets are conducted on randomly sampled subsets while preserving the original class distribution.

TABLE I. DATASETS DESCRIPTION

| Dataset | *Samples* | *Features* | *Classes* |
|---|---|---|---|
| **HIGGS** | 1M/11Million | 28 | 2 |
| **KDD Cup 1999** | 0.6M/4Million | 41 | 2 |
| **FARS** | 100,968 | 29 | 8 |
| **Connect-4** | 67,557 | 42 | 3 |
| **Synthetic** | 1Million | 50 | 10 |

We compare ASSS with several representative data reduction baselines, including Full Training, Random Subsampling, K-means Clustering Subsampling (MiniBatch centroids), GradientProxy (gradient-norm scoring), Herding (class-conditional diversity), and MRMC (entropy-based curriculum).

Evaluation metrics are designed both for predictive performance and data efficiency. Predictive quality is measured by AUC on the test sets, known as a robust indicator for classification and class imbalance. To quantify how well the reduced dataset preserves predictive information, we report the Performance Retention Rate (PRR), defined as the ratio between the performance achieved using the selected subset and that obtained training on full dataset. High PRR values indicate better preservation of task-relevant information under data compression.

For each dataset, standard scaling and 5-fold stratified cross validation are applied and 10% of the training data is reserved as a validation set for ASSS hyperparameter tuning. We set compression ratio $r = 0.3$ unless specified and all compared methods use the same classifier architecture (a 3-layer MLP) for final training to ensure fairness. For ASSS, both the selector network and the task network are 3-layer MLPs with hidden dimention 128. They are initialized using He-Uniform initialization. We use the Adam optimizer with default parameters ( $\beta_1 = 0.9, \beta_2 = 0.999$ ) for both networks. The learning rate for the task network is fixed at $10^{-3}$, the selector's learning rate is tuned via Bayesian optimization in Stage 2 (typically $10^{-4} \sim 5\times10^{-4}$ ). No explicit learning rate decay is applied during the adversarial phase, instead, we anneal the Gumbel-Softmax temperature τ exponentially from an initial value (optimized in Stage 2) down to 0.1. For final fine-tuning (Stage 5), we reduce the task learning rate to $10^{-4}$ and use a cosine decay schedule over 30 epochs.

### B. Results and Analysis

Table II summarises the mean AUC and PRR (with standard deviations) across all datasets. ASSS achieves the highest AUC among all subset-selection methods, corresponding to a PRR of 98.9% relative to training on the full dataset. This indicates that ASSS retains nearly all task-relevant information while using only 30% of the data. In contrast, traditional task-agnostic methods: random sampling, K-means centroid selection, gradient-norm based scoring, and MRMC all yield PRRs between 98.0% and 98.3%, while Herding lags behind with a PRR of only 97.0%. The standard deviations are comparable across methods, confirming the statistical significance of ASSS's advantage.

TABLE II. OVERALL PERFORMANCE

| Method | *AUC* | *PRR* |
|---|---|---|
| **Full** | 0.9803 ± 0.0033 | 100% |
| **ASSS** | 0.9693 ± 0.0044 | 98.9% |
| **GradientProxy** | 0.9635 ± 0.0038 | 98.3% |
| **K-Means** | 0.9636 ± 0.0038 | 98.3% |
| **Random** | 0.9626 ± 0.0046 | 98.2% |
| **MRMC** | 0.9610 ± 0.0044 | 98.0% |
| **Herding** | 0.9510 ± 0.0053 | 97.0% |

The t-SNE visualizations of a toy data using different methods in Fig. 3 provide direct evidence of ASSS's task-aware selection strategy. Subfigures are labeled (a) ASSS, (b) Random, and (c) K-Means. In (a), the retained samples form a balanced, well-separated distribution that preserves inter-class boundaries and rare discriminative configurations.In contrast, Random selection (b) produces dense, overlapping central masses with many redundant interior points, while K-Means (c) overly concentrates around cluster centroids, discarding boundary samples critical for decision-making. discarding boundary samples that are critical for decision-making. This visual pattern is consistent across all four datasets and directly explains ASSS's superior PRR: it preferentially retains support vectors and rare but informative instances rather than geometric prototypes.

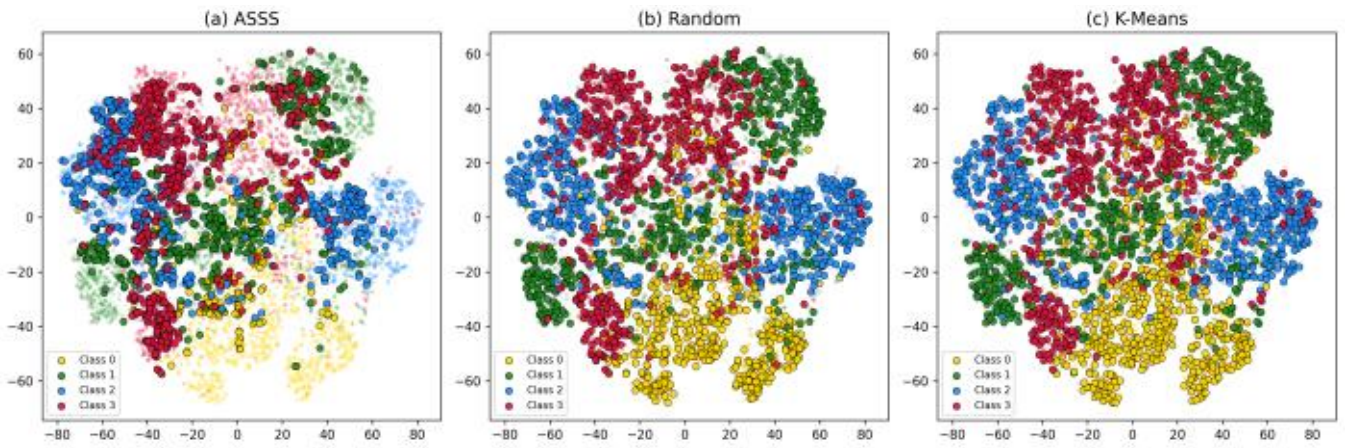

Fig. 3. T-SNE Visualization of Selected Samples. (a) ASSS retains boundary samples; (b) Random selects many interior points; (c) K-Means concentrates on cluster centroids.

### C. Summary

The experimental results demonstrate that ASSS consistently outperforms existing data reduction methods in terms of AUC and PRR. By casting subsampling as a differentiable adversarial game guided by the information bottleneck principle, ASSS learns to select a compact yet highly informative subset that preserves task-critical patterns. The t-SNE visualisations further validate that the selected points are not merely representative in a geometric sense, but are explicitly optimised for the downstream classification task.

## V. Conclusion

In this paper, we introduced ASSS, a novel differentiable adversarial framework for task-aware data subsampling. By casting subset selection as a minimax game between a selector network and a task network, and by using the Gumbel-Softmax reparameterisation for end-to-end differentiability, ASSS learns to retain the most informative samples for a given downstream objective. The method is underpinned by the information bottleneck principle, which provides a theoretical justification for the trade-off between compression and predictive accuracy.

Our experimental evaluation on a diverse set of datasets demonstrates that ASSS consistently outperforms existing data reduction techniques. With a compression ratio of 30%, ASSS achieves a performance retention rate of 98.9% in terms of AUC, approaching the accuracy of models trained on the full dataset. In contrast, traditional task-agnostic methods (random sampling, K-means, gradient-norm scoring, MRMC) attain PRRs between 98.0% and 98.3%, while Herding lags at only 97.0%. The t-SNE visualisations further reveal that ASSS preserves critical boundary samples and rare discriminative patterns, whereas conventional methods tend to select redundant interior points or cluster centroids, discarding the very instances that define decision boundaries. These results confirm that task-aware selection, as implemented by ASSS, is essential for achieving high performance under strong data compression.

Future work includes extending ASSS to complex task networks, refining the information-theoretic objective with tighter bounds, designing more expressive selectors, and applying the framework to reinforcement learning and continual learning. Theoretical analysis of the adversarial game's convergence also merits further investigation.

## Acknowledgment

This study was independently completed by the author without any external financial support or academic assistance. We express our gratitude to the academic community for research foundation and open academic environment.

## References

[1] Raza, Hassan, A. Singh, Tsendayush Erdenetsogt, Muhammad Mohsin Kabeer, Muhammad Shahrukh Aslam, and Mazhar Farooq. "Machine Learning Driven Decision Making in the Modern Data Era." PERFECT: Journal of Smart Algorithms 3, no. 1 (2026): 11-22.

[2] Shen, Li, Yan Sun, Zhiyuan Yu, Liang Ding, Xinmei Tian, and Dacheng Tao. "On efficient training of large-scale deep learning models." ACM Computing Surveys 57, no. 3 (2024): 1-36.

[3] Elmar, Kashyap Chitta Jose M. Alvarez, and Haussmann Clement Farabet. "Less is more: An exploration of data redundancy with active dataset subsampling." arXiv preprint arXiv:1905.12737 (2019).

[4] Wang, Jing, Jiahui Zou, and HaiYing Wang. "Sampling with replacement vs Poisson sampling: a comparative study in optimal subsampling." IEEE Transactions on Information Theory 68, no. 10 (2022): 6605-6630.

[5] Shang, Boyang, Daniel W. Apley, and Sanjay Mehrotra. "Diversity subsampling: Custom subsamples from large data sets." INFORMS Journal on Data Science 2, no. 2 (2023): 161-182.

[6] Dempsey, Walter. "Recurrent event analysis in the presence of real-time high frequency data via random subsampling." Journal of Computational and Graphical Statistics 33, no. 2 (2024): 525-537.

[7] Axiotis, Kyriakos, Vincent Cohen-Addad, Monika Henzinger, Sammy Jerome, Vahab Mirrokni, David Saulpic, David Woodruff, and Michael Wunder. "Data-efficient learning via clustering-based sensitivity sampling: Foundation models and beyond." arXiv preprint arXiv:2402.17327 (2024).

[8] Ruetz, Simon. "Adapted variable density subsampling for compressed sensing." Constructive Approximation 61, no. 3 (2025): 511-534.

[9] Yu, Jun, Mingyao Ai, and Zhiqiang Ye. "A review on design inspired subsampling for big data." Statistical Papers 65, no. 2 (2024): 467-510.

[10] Yao, Yaqiong, and HaiYing Wang. "A review on optimal subsampling methods for massive datasets." Journal of Data Science 19, no. 1 (2021): 151-172.

[11] Levina, Anna, Viola Priesemann, and Johannes Zierenberg. "Tackling the subsampling problem to infer collective properties from limited data." Nature Reviews Physics 4, no. 12 (2022): 770-784.

[12] Lyu, Jiacheng, Jie Yang, Zhixun Su, and Zilu Zhu. "Ld-smote: a novel local density estimation-based oversampling method for imbalanced datasets." Symmetry 17, no. 2 (2025): 160.

[13] Rodriguez-Diaz, Paula, Lingkai Kong, Kai Wang, David Alvarez-Melis, and Milind Tambe. "What is the right notion of distance between predict-then-optimize tasks?." arXiv preprint arXiv:2409.06997 (2024).

[14] Li, Li-li, Heng Xiao, Yu Wang, Haolun Shi, and Jiguo Cao. "OSK: Optimal Subsampling Method Based on K-means Clustering for Imbalanced Big Data." (2023).

[15] Zhu, Rong. "Gradient-based sampling: An adaptive importance sampling for least-squares." Advances in neural information processing systems 29 (2016).

[16] Bachem, Olivier, Mario Lucic, and Andreas Krause. "Practical coreset constructions for machine learning." arXiv preprint arXiv:1703.06476 (2017).

[17] Killamsetty, Krishnateja, Durga Sivasubramanian, Ganesh Ramakrishnan, and Rishabh Iyer. "Glister: Generalization based data subset selection for efficient and robust learning." In Proceedings of the AAAI conference on artificial intelligence, vol. 35, no. 9, pp. 8110-8118. 2021.

[18] Killamsetty, Krishnateja, Sivasubramanian Durga, Ganesh Ramakrishnan, Abir De, and Rishabh Iyer. "Grad-match: Gradient matching based data subset selection for efficient deep model training." In International Conference on Machine Learning, pp. 5464-5474. PMLR, 2021.

[19] Chen, Yueqi, Witold Pedrycz, Chao Zhang, Jian Wang, and Jie Yang. "Oversampling with GAN via meta-learning for imbalanced data." IEEE Transactions on Multimedia (2025).

[20] Jang, Eric, Shixiang Gu, and Ben Poole. "Categorical reparameterization with gumbel-softmax." arXiv preprint arXiv:1611.01144 (2016).

[21] Sachdeva, Noveen, and Julian McAuley. "Data distillation: A survey." arXiv preprint arXiv:2301.04272 (2023).

[22] Ren, Pengzhen, Yun Xiao, Xiaojun Chang, Po-Yao Huang, Zhihui Li, Brij B. Gupta, Xiaojiang Chen, and Xin Wang. "A survey of deep active learning." ACM computing surveys (CSUR) 54, no. 9 (2021): 1-40.

[23] Tishby, Naftali, Fernando C. Pereira, and William Bialek. "The information bottleneck method." arXiv preprint physics/0004057 (2000).

[24] Kirsch, Andreas. "Advancing deep active learning & data subset selection: Unifying principles with information-theory intuitions." arXiv preprint arXiv:2401.04305 (2024).

[25] M. Heusel, H. Ramsauer, T. Unterthiner, B. Nessler, and S. Hochreiter, "GANs trained by a two time-scale update rule converge to a local nash equilibrium," in Advances in Neural Information Processing Systems (NeurIPS), 2017, pp. 6626–6637.

[26] M. Kelly, R. Longjohn, and K. Nottingham, "The UCI Machine Learning Repository," 2017. [Online]. Available: https://archive.ics.uci.edu